# Structure Tensor Based Image Interpolation Method

Ahmadreza Baghaie and Zeyun Yu
University of Wisconsin-Milwaukee, WI, USA

**Abstract —** Feature preserving image interpolation is an active area in image processing field. In this paper a new direct edge directed image super-resolution algorithm based on structure tensors is proposed. Using an isotropic Gaussian filter, the structure tensor at each pixel of the input image is computed and the pixels are classified to three distinct classes; uniform region, corners and edges, according to the eigenvalues of the structure tensor. Due to application of the isotropic Gaussian filter, the classification is robust to noise presented in image. Based on the tangent eigenvector of the structure tensor, the edge direction is determined and used for interpolation along the edges. In comparison to some previous edge directed image interpolation methods, the proposed method achieves higher quality in both subjective and objective aspects. Also the proposed method outperforms previous methods in case of noisy and JPEG compressed images. Furthermore, without the need for optimization in the process, the algorithm can achieve higher speed[1].

## 1. Introduction

Feature preserving image interpolation is an active area in the image processing field, from everyday digital pictures to application-oriented medical and satellite images. Many methods have been proposed in the past decades to tackle this problem [1-20]. Generally speaking, the methods for image interpolation/super-resolution can be divided in 3 different categories: 1) Direct Interpolation methods, 2) PDE based interpolation methods and 3) Optimization based interpolation methods. All of these methods have their pros and cons regarding their simplicity of implementation, computational complexity and performance. The proposed method in this paper is a direct interpolation method without the need for any optimization in the process. Also in terms of computational time, the proposed method can achieve the result in less than one second for an image of typical size using MEX based implementation. This feature along with not using any optimization procedure, as well as being robust in case of noisy images, make this method a suitable choice for implementation in everyday used electronic devices.

Nearest neighbor and bilinear interpolation are two simple methods for image interpolation [1]. Despite the simplicity in implementation and very low computational cost, these methods suffer from severe blocky artifacts, as well as blurring and ringing artifacts near the edges. Although better performance can be achieved by using higher order splines, rather than 0 and 1 order splines as in the nearest neighbor and bilinear methods, higher order spline methods still contain oscillatory edges and ringing artifacts [2]. The main reason is that these methods don't take into consideration any information other than intensity values. In other words, they are intensity based and not feature (edge) based. So even though they are easy to implement and need low computational cost, they are not suitable for most of applications.

The final recipient of any image processing algorithm is the human visual system which is very feature sensitive. These features are mostly edges and corners within the image. Also sharpness of the final image is of high importance. Based on these criteria, the previously mentioned methods, despite their technical advantages, are not satisfactory. So the need for introducing new approaches and novel models for image interpolation which satisfy the human visual system has been emerged in the past decades and many methods have been proposed. Some of these methods will be mentioned here.

[1] Ahmadreza Baghaie (Corresponding author) is with the Department of Electrical Engineering at University of Wisconsin - Milwaukee, Milwaukee, WI 53211 (abaghaie@uwm.edu).
Zeyun Yu is with the Department of Computer Science at University of Wisconsin - Milwaukee, Milwaukee, WI 53211 (yuz@uwm.edu).

Edge directed methods usually are the first ones that come to notice when dealing with image interpolation problem. In 2001 a method called NEDI was proposed which performs based on the assumption that every image can be modeled as a locally stationary Gaussian process [3]. Based on this assumption, the local covariance coefficients from the low resolution (LR) image is estimated and then interpolation is done based on the geometric duality between the LR covariance and the high resolution (HR) covariance. An improved version of NEDI algorithm called iNEDI is proposed later which achieves higher scores in terms of subjective and objective image quality measures relative to NEDI with the cost of needing more computational time [4]. Another edge directed image interpolation method is ICBI which works based on an estimation of the edge orientation using second order derivative of the image [5]. DFDF method [6] is another method in this category which utilizes directional filtering and data fusion. In DFDF, at first, two observation sets are defined in two orthogonal directions for each pixel to be interpolated. Then these two estimates will be fused using Linear Minimum Mean Square Error (LMMSE) in order to achieve a more robust estimate for the missing pixel.

Methods proposed in [7-15] also are good examples of edge directed image interpolation. In [7], the method is based on partitioning the input image into homogeneous and edge areas with regard to local structure of the image and then, interpolating each parts differently, bilinear interpolation for homogeneous regions and an adaptive edge oriented method for edge pixels. In [8], a modified edge adaptive bilinear image interpolation method called EASE is proposed. This modified version is achieved using the classical interpolation error theorem. In [9], a new directional cubic convolution (CC) interpolation scheme is proposed. In [10], an interpolation framework is proposed in which denoising and image sharpening are embedded together. In this method, bilateral filtering method is used to partition the input image into detail and base layers, and then edge preserving interpolation method is applied to each layer. In [11], the edge information of the LR image is first estimated using the modified Leung-Malik filter bank, and then this information is converted into that of HR image by using a mapping function. In [12], a fast image interpolation method with adaptive weights is proposed motivated by Inverse Distance Weighting (IDW). The use of Radial Basis Functions (RBFs) to solve image interpolation problem is investigated in [13, 14]. In [15], a soft decision interpolation technique is proposed which estimates missing pixels in groups rather than one at a time. They use a piecewise 2-D autoregressive (AR) model to determine the local structure of the scene.

Even though the above mentioned methods are of a wide range of use and discipline, still there are more methods that are not discussed here; like Partial Differential Equation (PDE) based methods [16, 17, 27], and regularization based methods [18-20]. The reader will be referred to the papers and their references for more information on these classes of image interpolation methods.

As can be seen, each of the mentioned methods deals with the image interpolation problem from a different angle. But still image interpolation is an open problem and there is room for improvement. In this paper, a new edge-directed method based on structure tensor will be proposed which its strength is not only in reconstructing edges in the HR image, but also is more robust in case of noise. The proposed method is very simple and easy to implement and based on the conducted experiments, outperforms the most common image interpolation methods. For comparison, five well-known image interpolation methods are considered: NEDI [3], DFDF [5], ICBI [6], KR [26] and iNEDI [4]. Tests were conducted for noise-free, noisy and JPEG compressed images. For completeness of the comparison another structure tensor-based method by Roussos and Maragos [27] is also considered. This method (RM) can be categorized as a PDE-based technique. In this method at first an initial interpolation is done by Fourier zero-padding and de-convolution. The result of this stage suffers from significant ringing artifacts. Using a tensor-driven diffusion process, the ringing artifacts are removed. The main assumption in this method is that the process of interpolation is a reversible process which cannot be hold always. Based on this assumption, interpolation is done by first applying an anti-aliasing low-pass filter followed by sampling. This assumption can be problematic especially in the case of naïve sub-sampling which is the case used in this paper. In naïve sub-sampling of factor $N$, one pixel is chosen from each $N$ pixels of the image without any anti-aliasing filtering. This will cause high amount of ringing artifacts near edges introduced by the first stage of the method as well as sever stair-cased edges which cannot be resolved properly using the tensor-driven diffusion process. More discussion will be given in the following sections regarding this issue. Here the online implementation of this method is used implemented by Getreuer [28].

The rest of the paper is organized as follows: in Section 2 a brief introduction will be given on structure tensor computation and its theoretical aspects. Then in Section 3, the proposed interpolation method will be described in more detail. Section 4 contains the implementation aspects, image quality measures that being used and tables of objective and subjective comparison between the five above mentioned methods and the proposed method, as well as some of the final results. Section 5 concludes the paper.

## 2. LOCAL STRUCTURE TENSOR

Local structure tensors have been used in image processing to solve problems such as anisotropic filtering [21, 22] and motion detection [23]. This method uses the gradient information of an image in order to determine the orientation information of the edges and corners. The structure tensor is defined as:

$$T_\sigma = \begin{bmatrix} g_x^2 * G_\sigma & g_x g_y * G_\sigma \\ g_y g_x * G_\sigma & g_y^2 * G_\sigma \end{bmatrix} = \begin{bmatrix} T_{11} & T_{12} \\ T_{12} & T_{22} \end{bmatrix} \quad (1)$$

where $G_\sigma$ is a Gaussian function with standard deviation $\sigma$, and$g_x$ and $g_y$are horizontal and vertical components of the gradient vector at each pixel respectively. Since matrix $T_\sigma$ is symmetric and positive semi-definite, it has two orthogonal eigenvectors as follows:

$$V = \begin{pmatrix} T_{22} - T_{11} + \sqrt{(T_{22} - T_{11})^2 + 4T_{12}^2} \\ -2T_{12} \end{pmatrix}, \quad \text{and normalized as: } V = \frac{V}{\|V\|} \quad (2)$$

$$V^\perp = \begin{pmatrix} 2T_{12} \\ T_{22} - T_{11} + \sqrt{(T_{22} - T_{11})^2 + 4T_{12}^2} \end{pmatrix}, \quad \text{and normalized as: } V^\perp = \frac{V^\perp}{\|V^\perp\|} \quad (3)$$

The corresponding eigenvalues for each eigenvector are as follows:

$$d = \frac{1}{2}\left(T_{22} + T_{11} - \sqrt{(T_{22} - T_{11})^2 + 4T_{12}^2}\right) \quad (4)$$

$$d^\perp = \frac{1}{2}\left(T_{22} + T_{11} + \sqrt{(T_{22} - T_{11})^2 + 4T_{12}^2}\right) \quad (5)$$

Apparently the eigenvalue $d$ is smaller than $d^\perp$. Based on the two eigenvalues, local structures can be determined as one of three types:

- Constant areas: $d^{\perp} \approx d \approx 0$
- Edges: $d^{\perp} \gg d \approx 0$
- Corners: $d^{\perp} \approx d \gg 0$

For edge points, the eigenvector $V$corresponding to the smaller eigenvalue is along the edge (tangent direction), while the eigenvector $V^{\perp}$is across the edge (normal direction).

Although using gradient vectors in an image can determine the edge orientations too, there are some other advantages in using structure tensors compared to gradient vectors alone. First, the edges in an image may not be smooth and continuous, especially in down-sampled images. With the Gaussian filtering of the gradient vectors in a neighborhood, as seen in the definition of the structure tensor, one can acquire more robust and accurate estimation of edge orientations. Second, the structure tensor can classify local features into several distinctive types, which is nontrivial by using gradient vectors alone. This becomes more obvious when a three-dimensional image or cloud of points is being considered. Also because of the Gaussian filtering stage, the edge orientation achieved by structure tensor is more robust against noise.

## 3. Methodology

### *3.1 Structure Tensor Based Image Interpolation*

Without loss of generality, only doubling the size of input image is considered. The same approach can be used for other scaling factors. Consider $I_{LR}$ as input image with size of $M \times N$ which is to be enlarged by a factor of 2 in both directions to produce $I_{HR}$ with size of $(2M-1) \times (2N-1)$. Illustrations of $I_{LR}$ and $I_{HR}$ pixels can be seen in Fig.1 where dark pixels represent the pixels from $I_{LR}$ and white pixels are those added to generate $I_{HR}$. In this case:

$$I_{HR}(2m-1,2n-1) = I_{LR}(m,n) \quad for\colon\ m \in [1,M], n \in [1,N] \qquad (6)$$

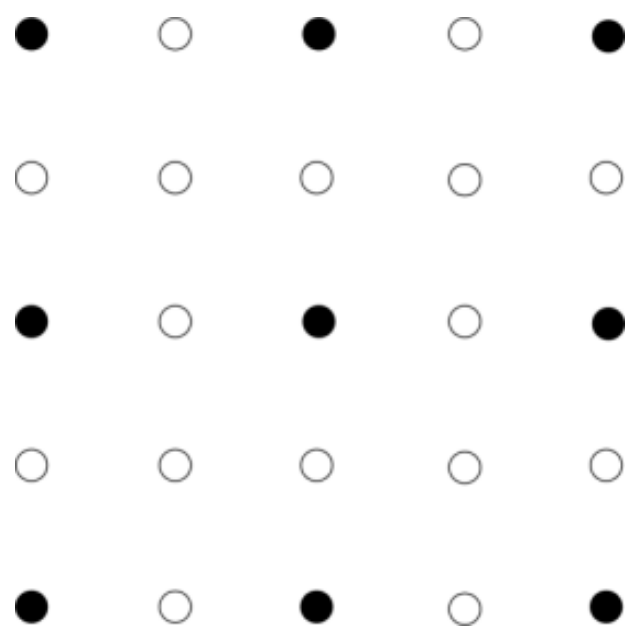

**Fig.1 Configuration of the $I_{LR}$ and $I_{HR}$ pixels**

After computing the structure tensor for the LR image, the edge orientation for each pixel of the input image is obtained. The remaining task is to compute the intensity values for the new pixels (white in Fig. 1) in the interpolated image. Assume the pixel to be interpolated is located at $(m_s, n_s)$ where $1 < m_s < M, 1 < n_s < N$ (see Fig.2).

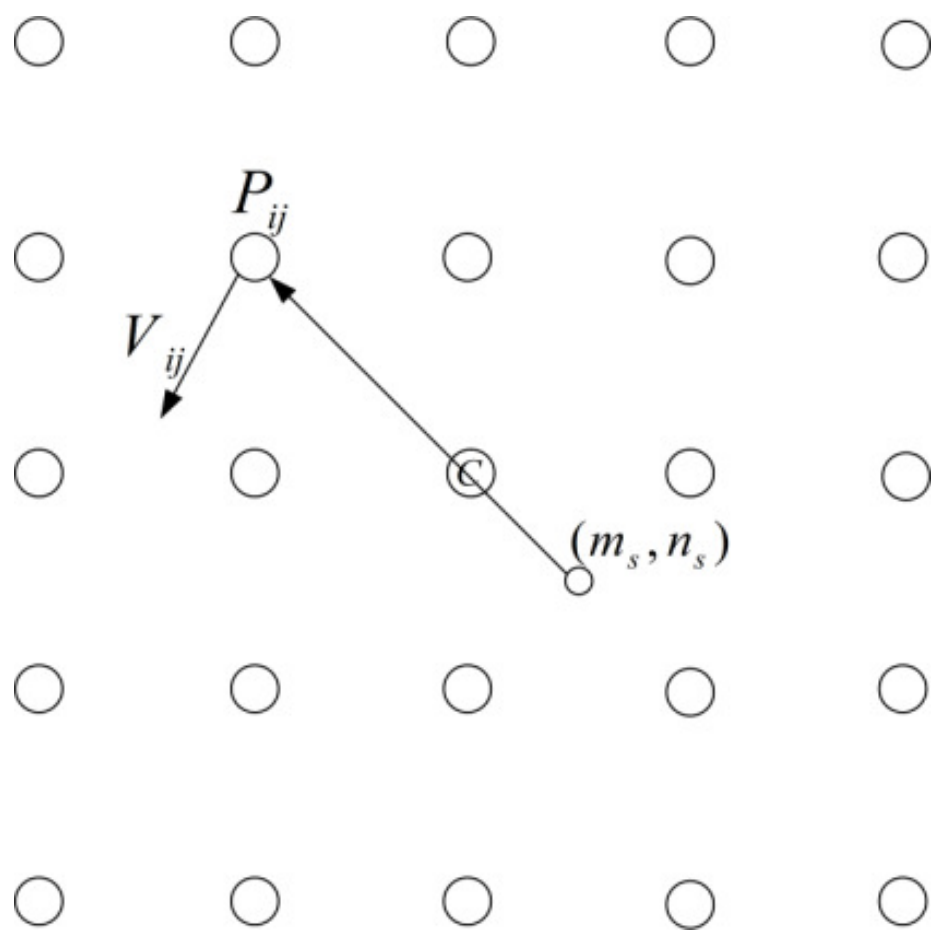

**Fig.2 Configuration of the new pixel at location $(m_s, n_s)$ w.r.t known pixels from input image**

The intensity value for the pixel to be interpolated is defined as a weighted summation of pixels in a defined neighborhood. Here a square neighborhood for averaging is defined. The complete form of the weighted average is:

$$I_{HR}(2m_s - 1, 2n_s - 1) = \sum_{i=\lfloor m_s \rfloor - D}^{\lfloor m_s \rfloor + D} \sum_{j=\lfloor n_s \rfloor - D}^{\lfloor n_s \rfloor + D} W_d(i,j) W_T(i,j) I_{LR}(i,j) \qquad (7)$$

where D is half of the neighborhood size and $W_d$ and $W_T$ are weight functions. $W_d$ is the distance based part of the weight function and it can be defined for pixel (i,j) in the neighborhood as follows:

$$W_d(i,j) = e^{-\beta(\|C - P_{ij}\|)} \qquad (8)$$

where C is the location of nearest pixel to $(m_s, n_s)$: $C = (\lfloor m_s \rfloor, \lfloor n_s \rfloor)$ and $P_{ij}$ is the location of pixel (i,j) in the neighborhood (see Fig.2). $W_T$ is the structure tensor based part of the weight function.

As previously mentioned, the eigenvalues and eigenvectors of the structure tensor at a pixel can be used to determine the tangent and normal directions at the pixel in the input image. To reduce the staircase artifact in the interpolated image, the interpolation should be performed along the edges. For this reason, the tangent eigenvector is used as a measure for computing the weight $W_T$. As shown in Fig.2 where the pixel at $(m_s, n_s)$ needs to be interpolated, for every pixel with known intensity value in the neighborhood, a vector connecting $(m_s, n_s)$ to the pixel can be defined. The corresponding weight is defined in such a way that only pixels with similar edge direction as the connecting vector should be assigned higher weights. In other words, even though the defined neighborhood is isotropic, the shape of the structure tensor based weight is not symmetric, unlike the distance based weight function. Therefore $W_T$ is defined as follows:

$$W_T(i,j) = e^{\gamma \left| dot\left(V_{ij}, \frac{(C - P_{ij})}{\|C - P_{ij}\|}\right) \right|} \qquad (9)$$

Where $V_{ij}$ is the tangent eigenvector of the pixel at (i, j) in the neighborhood and the *dot* denotes the dot product of two normalized input vectors.

Using this formulation for computing the total weight, not only the distance between the pixel to be interpolated and its neighboring pixels but also the edge orientation of the neighboring pixels are taken into consideration. The main difference between this method and other gradient based methods is that the edge orientation is achieved using structure tensors and thus the blocky artifacts are significantly reduced.

### *3.2 Implementation*

Apparently a straightforward implementation of the proposed algorithm can be very time consuming. For example, assuming a 5x5 neighborhood size for each new pixel to be interpolated, 25 weights for each of the distance based and structure tensor based weights should be computed before the computation of the weighted summation. Fortunately, in most digitized images, only a small portion of pixels are edge/corner regions leaving a large number of pixels in uniform regions with very small variations in gray values. These regions don't contain as much important information as edges/corners and hence they can be easily and efficiently interpolated using simple and fast interpolation methods like bilinear interpolation. To do this, a very simple pair of gradient masks is implemented. Fig.3 shows the gradient masks:

| | | |
|---|---|---|
| 1 | 0 | -1 |
| 1 | 0 | -1 |
| 1 | 0 | -1 |
| 1 | 1 | 1 |
| 0 | 0 | 0 |
| -1 | -1 | -1 |

**Fig.3 Gradient masks (left:$g_x$; right: $g_y$)**

Convolving these masks with the input image, we can compute the gradient values for each pixel in horizontal and vertical directions and the magnitude of the gradient is computed using $\boldsymbol{g} = \sqrt{(g_x^2 + g_y^2)}$ for each pixel. After normalizing the magnitude into the range [0,100], a simple threshold ($\boldsymbol{T}$) is applied in order to filter out the pixels in relatively uniform regions, where the pixels will be interpolated with a simple and fast interpolation method.

Another issue rises in dealing with corner points. When computing the structure tensor, the Gaussian filter tends to smooth and round the corners. To that end, the corner points should be treated differently than edge points. Using the structure tensor, identification of corner points is easy: not only the smaller eigenvalue of the structure tensor significantly is bigger than 0, but also the ratio of the smaller and bigger eigenvalues is greater than that for edge pixels. Using this criterion, corner points can be detected.

Based on the above considerations, the proposed algorithm is given below.

**ALGORITHM 1: STRUCTURE TENSOR BASED IMAGE INTERPOLATION (STB)**

**Structure Tensor Based Image Interpolation (STB)**
**Inputs**:
**$I_{LR}$, D, β, γ, σ, T**
***Preprocessing***:
Gradient computation and edge regions detection:
**$g_x$, $g_y$:** Gradient in x and y direction
**$g_{mag}$:** Normalized magnitude of the gradient in range [0, 100]
**$I_E$:** Image's edge map, using threshold **T**
Structure tensor computation:
Defining the Gaussian filter using **σ**;

```
  Computing d, d⊥, V, V⊥
Interpolation:
        For i=D+1 to M-D with step=1/2
              For j=D +1 to N-D with step=1/2
                      If ⌊i⌋ = i and ⌊j⌋ = j
                            I_HR(2i − 1, 2j − 1) = I_LR(i, j);
                      Else if C = (⌊m_s⌋, ⌊n_s⌋) is in uniform region or is a corner point
                            Bilinear Interpolation;
                      Else
                            I_HR(2m_s − 1, 2n_s − 1) = Σ_{i=⌊m_s⌋−D}^{⌊m_s⌋+D} Σ_{j=⌊n_s⌋−D}^{⌊n_s⌋+D} W_d(i, j) W_T(i, j) I_LR(i, j)
                      End
              End
        End
```

## 4 Experimental Results

### 4.1 Image Quality Measures

In order to assess the proposed algorithm, several image quality measures were used. The most popular measure is Peak Signal to Noise Ratio (PSNR) which measures the intensity differences between two images [24]. Assume that X is the original image, and Y is the reconstructed image from the downsampled version. The Mean Squared Error (MSE) between X and Y is defined as follows:

$$MSE = \frac{1}{N}\sum_{i=1}^{N}(x_i - y_i)^2 \qquad (10)$$

Where $x_i$ and $y_i$ are the i[th] pixel of the original and reconstructed image respectively and N is the total number of pixels. Based on MSE, PSNR is defined as follows:

$$PSNR = 10\ log_{10}\frac{L^2}{MSE} \qquad (11)$$

where L is the dynamic range of pixel intensities in the images.

Another measure that is used is Edge PSNR (EPSNR) which is defined in the same manner as above, but instead uses the edge maps of the original and reconstructed images. For edge map computation, a simple Sobel operator is used.

Even though PSNR and EPSNR are proper measures for image quality comparison, they are objective and usually fail in describing the visual perception of images. To remedy this problem, several subjective image quality measures were proposed in literature. A well-known measure is the Structural SIMilarity(SSIM) index [24]. Recently another method called Feature SIMilarity (FSIM) index is proposed for image quality comparison [25]. In the following, these four measures are used to assess and compare the proposed method and several other interpolation techniques.

### 4.2 Results and Comparisons

In order to evaluate the performance of the proposed method, several images were tested. Fig.4 displays the test images considered in this paper. For each image, a direct downsampling procedure with a factor of 2 is performed in order to produce the $I_{LR}$ image. Then with the STB interpolation method the images were enlarged.

The performance of our algorithm is tested, using both noise-free and noisy images. For noisy images, compressed images are considered for comparison. JPEG format with 75% as quality is used here. Also for completeness of the experiments, our algorithm is also tested on images with added Gaussian noise. Tables 1-2 represent objective and subjective measures of the proposed method in comparison with several popular image interpolation methods for noise free images. For all of methods the default parameters are used. The default parameters for STB method are as follows: ***σ= 2, D=2, β=5, γ=10, T=20.***

As can be seen from Tables 1-2, the STB method performs very well in comparison with other methods. On the other hand, RM method performs poorly mainly because of the assumption of reversible interpolation. This assumption cannot be hold in case of naïve sub-sampling which is the case here. This is mainly due to the Fourier zero-padding and de-convolution in the first stage of the algorithm which causes significant ringing artifacts near edges. On some of the images, the iNEDI approach also works well. However, this method does not perform as well as our method when dealing with compressed images. Table 3 represents the objective and subjective comparisons between STB and iNEDI for compressed images.

For comparing the results of the proposed algorithm vs. iNEDI in case of noisy images, an additive Gaussian noise (zero mean with 0.1% variance) is applied to downsampled images, and then used our method and iNEDI to produce the enlarged images. Table 4 summarizes the results of objective and subjective image quality measures for test images.

As can be seen in Table 4, the STB outperforms iNEDI in almost all of the images with noticeable margin. Fig 5-6 show the overall results of different interpolation methods on some of the test images.

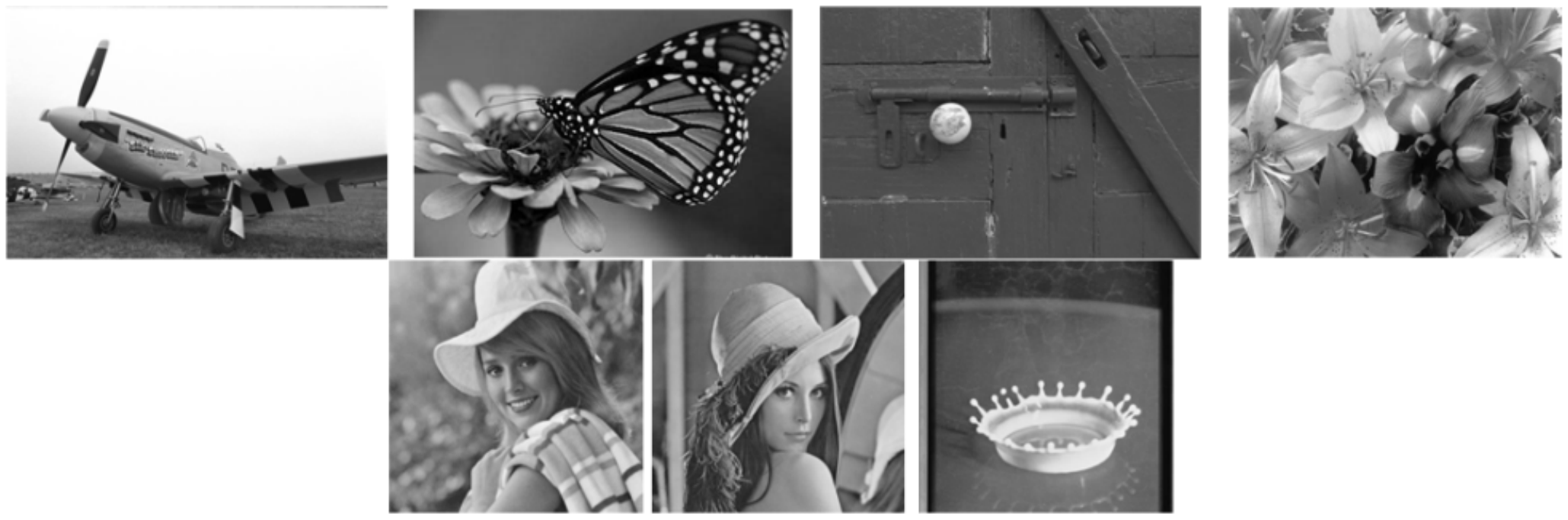

**Fig.4 Images used for comparison, from left to right, top to bottom: Airplane (512x768), Butterfly (324x492), Door (512x512), Flowers (480x640), Girl (512x512), Lena (512x512), Splash (512x768)**

**TABLE 1 :OBJECTIVE QUALITY COMPARISON OF DIFFERENT INTERPOLATION METHODS**

| Method | NEDI | | DFDF | | ICBI | | KR | | RM | | iNEDI | | STB | |
|---|---|---|---|---|---|---|---|---|---|---|---|---|---|---|
| | PSNR | EPSNR | PSNR | EPSNR | PSNR | EPSNR | PSNR | EPSNR | PSNR | EPSNR | PSNR | EPSNR | PSNR | EPSNR |
| **Airplane** | 28.69 | 15.42 | 30.53 | 19.44 | 30.08 | 18.68 | 29.11 | 16.01 | 27.64 | 18.28 | 30.66 | 19.65 | **30.71** | **19.82** |
| **Lena** | 33.57 | 27.75 | 33.96 | 28.10 | 34.05 | 26.99 | 33.97 | 27.97 | 30.41 | 24.89 | **34.11** | 27.86 | 33.99 | **28.70** |
| **Flowers** | 25.62 | 19.94 | 25.74 | 20.40 | 25.20 | 19.24 | 25.79 | 20.23 | 22.61 | 17.60 | 25.89 | 20.66 | **26.01** | **20.78** |
| **Girl** | 31.84 | 28.30 | 31.81 | 29.41 | 31.27 | 28.82 | 31.92 | 29.10 | 29.02 | 26.46 | **32.24** | 29.30 | 31.99 | **29.46** |
| **Door** | 32.14 | 25.73 | 32.27 | 25.93 | 31.71 | 24.33 | 32.20 | 25.86 | 28.56 | 22.42 | 32.41 | 25.99 | **32.47** | **26.25** |
| **Splash** | 31.38 | 15.49 | 33.79 | 19.79 | 33.32 | 21.29 | 33.38 | 19.07 | 32.45 | 20.55 | 33.69 | **21.28** | **35.23** | 21.26 |
| **Butterfly** | 28.96 | 21.02 | 29.67 | 20.76 | 29.91 | 20.26 | 29.54 | 21.25 | 27.15 | 18.79 | **30.07** | 21.36 | 29.31 | **21.49** |

TABLE 2: SUBJECTIVE QUALITY COMPARISON OF DIFFERENT INTERPOLATION METHODS

| Method | NEDI | | DFDF | | ICBI | | KR | | RM | | iNEDI | | STB | |
|---|---|---|---|---|---|---|---|---|---|---|---|---|---|---|
| | SSIM | FSIM | SSIM | FSIM | SSIM | FSIM | SSIM | FSIM | SSIM | FSIM | SSIM | FSIM | SSIM | FSIM |
| **Airplane** | .9110 | .9782 | .9144 | .9804 | .9085 | .9781 | .9049 | .9798 | .8606 | .9710 | .9166 | .9811 | **.9177** | **.9815** |
| **Lena** | .9112 | .9862 | .9129 | .9871 | .9112 | .9868 | .9097 | .9875 | .8448 | .9807 | **.9175** | **.9876** | .9147 | .9875 |
| **Flowers** | .6884 | .9412 | .6844 | .9390 | .6651 | .9316 | .6648 | .9440 | .5914 | 9143 | **.6974** | .9428 | .6940 | **.9450** |
| **Girl** | .7842 | .9684 | .7741 | .9656 | .7523 | .9590 | .7675 | .9701 | .7055 | .9464 | **.7933** | **.9717** | .7805 | .9698 |
| **Door** | .8576 | .9621 | .8607 | .9637 | .8501 | .9575 | .8501 | **.9633** | .7570 | .9506 | .8625 | .9622 | **.8662** | .9631 |
| **Splash** | .9285 | .9829 | .9296 | .9629 | .9241 | .9826 | .9245 | .9824 | .8677 | .9783 | **.9328** | **.9852** | .9313 | .9836 |
| **Butterfly** | .9427 | .9462 | .9501 | .9546 | **.9513** | **.9549** | .9468 | .9464 | .9099 | .9221 | .9507 | .9518 | .9475 | .9498 |

TABLE 3: OBJECTIVE AND SUBJECTIVE QUALITY COMPARISON FOR COMPRESSED IMAGES

| Method | iNEDI | | | | STB | | | |
|---|---|---|---|---|---|---|---|---|
| | PSNR | EPSNR | SSIM | FSIM | PSNR | EPSNR | SSIM | FSIM |
| **Airplane** | 29.84 | 19.34 | 0.8803 | 0.9716 | **30.50** | **19.76** | **0.9104** | **0.9804** |
| **Lena** | 32.54 | 26.73 | 0.8773 | 0.9789 | **33.78** | **28.49** | **0.9097** | **0.9892** |
| **Flowers** | 25.20 | **20.33** | 0.6362 | **0.9320** | **25.37** | 20.22 | **0.6729** | 0.9273 |
| **Girl** | 31.24 | 28.14 | 0.7468 | 0.9646 | **32.44** | **29.57** | **0.7931** | **0.9809** |
| **Door** | 31.26 | 25.24 | 0.8009 | 0.9493 | **32.28** | **26.19** | **0.8536** | **0.9643** |
| **Splash** | 33.38 | **20.04** | 0.8908 | 0.9758 | **33.56** | 19.87 | **0.9242** | **0.9824** |
| **Butterfly** | **29.09** | 21.01 | 0.9185 | 0.9277 | 29.04 | **21.22** | **0.9408** | **0.9433** |
| **Average** | 30.36 | 22.97 | 0.8215 | 0.9571 | **30.99** | **23.61** | **0.8578** | **0.9668** |

TABLE 4: OBJECTIVE AND SUBJECTIVE QUALITY COMPARISON FOR NOISY IMAGES

| Method | iNEDI | | | | STB | | | |
|---|---|---|---|---|---|---|---|---|
| | PSNR | EPSNR | SSIM | FSIM | PSNR | EPSNR | SSIM | FSIM |
| **Airplane** | 28.71 | 19.46 | 0.7638 | 0.9050 | **28.79** | **19.50** | **0.7660** | **0.9063** |
| **Lena** | 29.84 | 26.28 | 0.7409 | 0.9274 | **30.17** | **27.15** | **0.7469** | **0.9344** |
| **Flowers** | 24.74 | 20.22 | 0.6106 | 0.9036 | **25.10** | **20.44** | **0.6201** | **0.9159** |
| **Girl** | 28.90 | 27.29 | 0.6518 | 0.9175 | **29.24** | **27.76** | **0.6583** | **0.9256** |
| **Door** | 29.15 | 25.01 | 0.6631 | 0.8976 | **29.47** | **25.34** | **0.6760** | **0.9056** |
| **Splash** | 30.51 | 20.93 | 0.7108 | 0.9024 | **30.69** | **20.94** | **0.7206** | **0.9107** |
| **Butterfly** | **27.84** | 20.85 | 0.7767 | 0.8860 | 27.58 | **21.09** | **0.7793** | **0.8844** |
| **Average** | 28.53 | 22.87 | 0.7025 | 0.9056 | **28.73** | **23.18** | **0.7096** | **0.9118** |

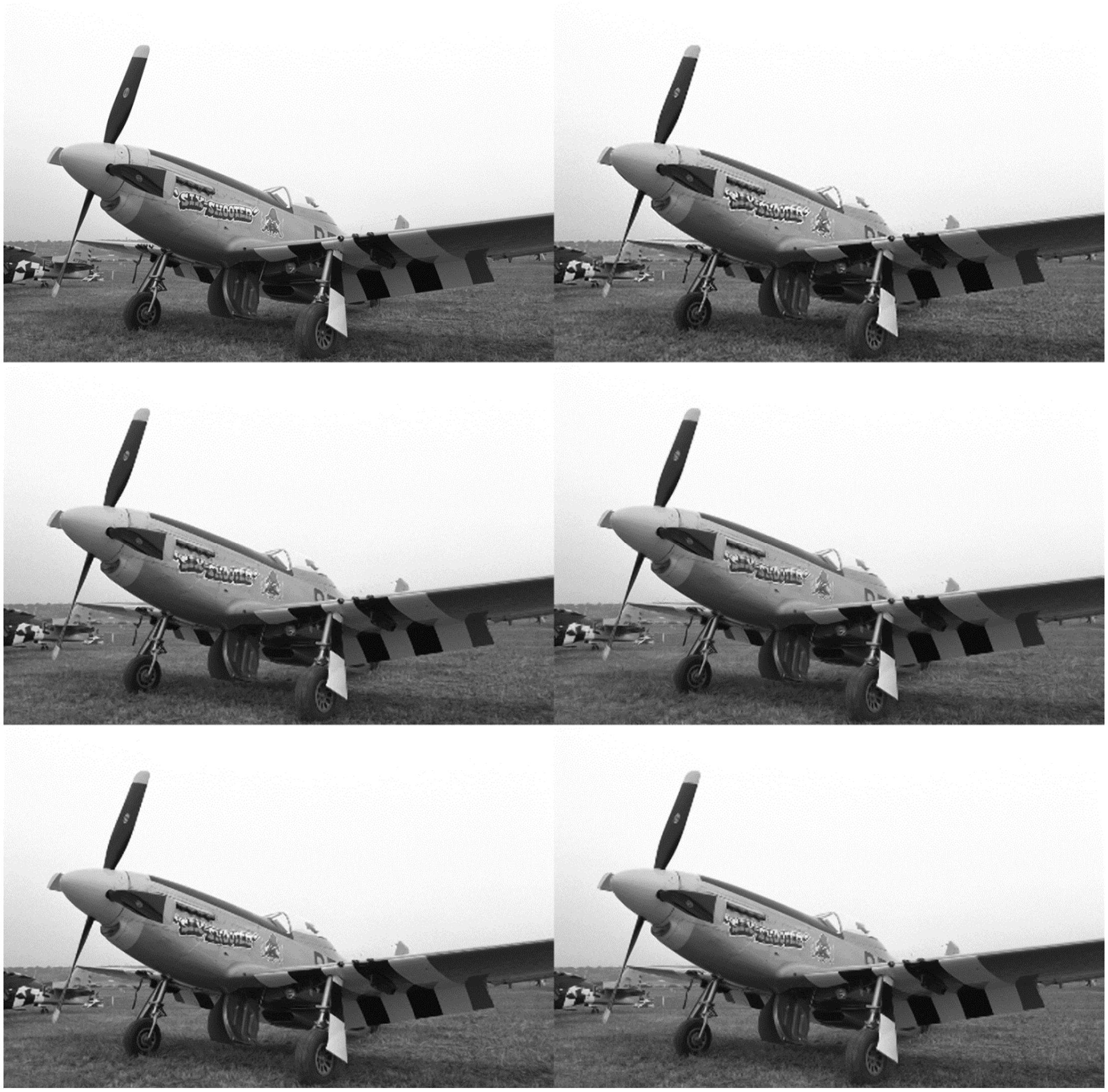

**Fig 5. Overall results of interpolation using different methods for Airplane. From left to right, top to bottom: Original image, ICBI, DFDF, NEDI, iNEDI, STB.**

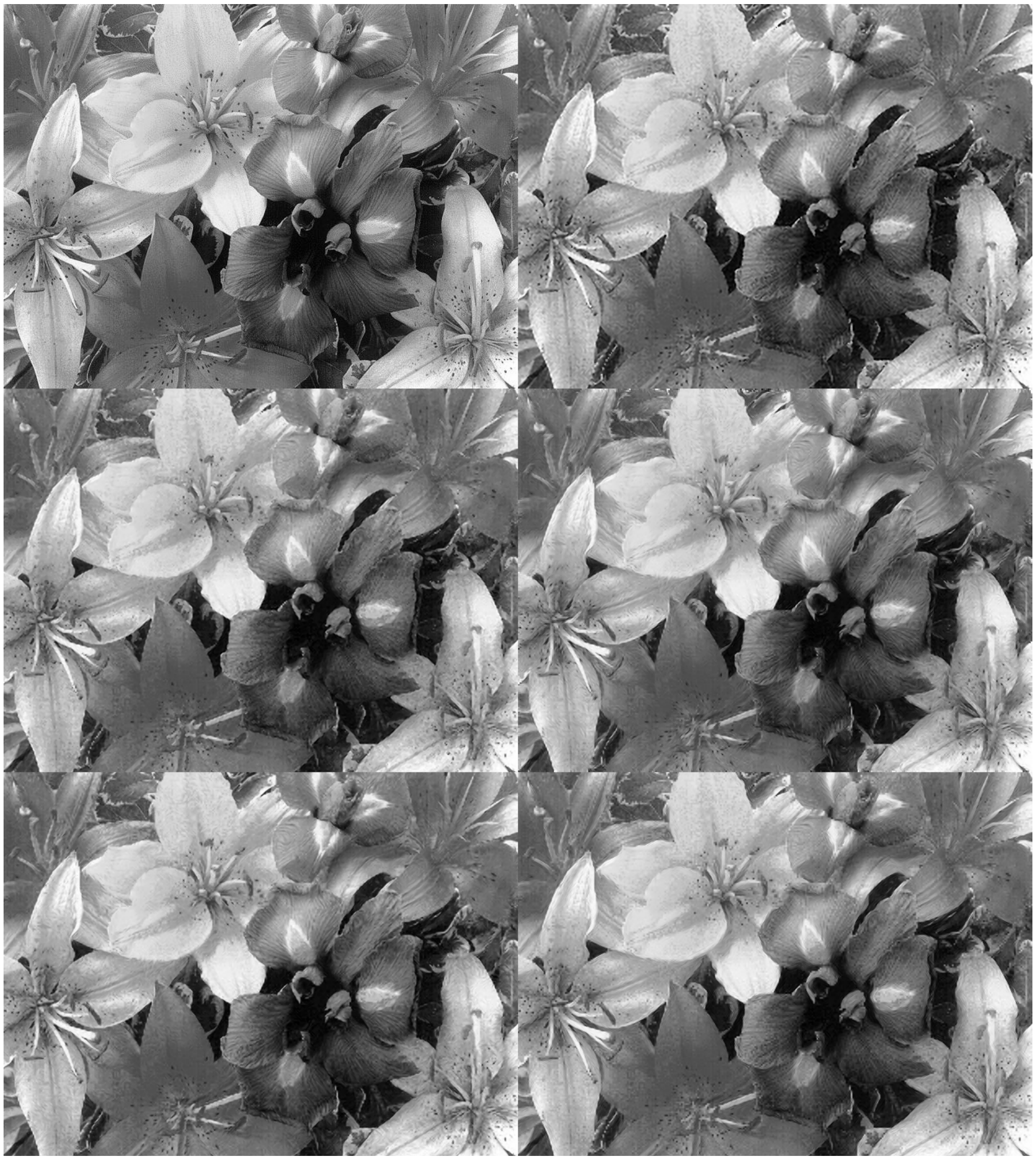

**Fig 6. Overall results of interpolation using different methods for Flowers. From left to right, top to bottom: Original image, ICBI, DFDF, NEDI, iNEDI, STB.**

The STB method performs the best in case of sharp edges. Fig. 7 visually shows a close view of the airplane's propeller generated by several methods including the proposed STB method.

As for the visual comparison of the two structure-tensor based methods, RM and STB, Fig. 8 shows the interpolation results as well as the differences between the results and original Lena image. Due to assumption of reversible interpolation which is not hold for the naïve sub-sampling, the RM results suffer from zig-zag edges from the first stage of the algorithm (Fourier zero-padding and deconvolution) that are not fully recovered by the tensor-driven diffusion process on the second stage. Fourier zero-padding and deconvolution causes severe ringing artifacts and zigzag edges. The diffusion process can resolve the ringing artifacts in the uniform areas, but is not able to fully recover the edges. Changing the parameters of the algorithm for further smoothing of the edges makes the uniform areas significantly smooth and eliminates the fine patterns.

Another important aspect of image interpolation methods is the computational cost. This has become more important when images are now digitized in much higher resolutions. Table 5 shows the average computational time for the test images using different methods. The RM method is excluded from this table since the available implementation is in C and not MATLAB. The tests were performed on a 3GHz Intel Core 2 Duo desktop with 4 GB of RAM. It can be seen that the proposed STB method is the fastest as compared to several other popular approaches. To further reduce the time cost, especially for some real-time applications, a MEX version of the proposed algorithm is implemented and the computational time was reduced to less than 1 second.

TABLE 5: COMPARISON OF AVERAGE COMPUTATIONAL TIME FOR DIFFERENT METHODS

| Method | NEDI | DFDF | ICBI | iNEDI | STB |
|---|---|---|---|---|---|
| Time (sec) | 21.5 | 19.6 | 127.4 | 800.2 | **11.4** |

## 4. CONCLUSION

In this paper, a new edge directed method for image interpolation based on structure is proposed which takes into account the advantages of structure tensor to determine the edge orientation as well as corner points of an image. Even though the concept of structure tensor is the same as gradient vectors, due to the presence of noise and discontinuity of edges caused by image compression, downsampling and digitizing, the structure tensor provides more robust edge orientations. Also using structure tensor, corner points can be better distinguished from other features such as edges.

The proposed method is tested for noise-free, noisy and JPEG compressed images. Numerous comparisons were made against several popular image interpolation methods. In most cases the proposed method outperformed the other methods both subjectively and objectively, especially in case of noisy and compressed images with noticeable margin. Also in terms of computational time, the proposed method can achieve the result in less than one second for an image of typical size using MEX based implementation. This feature along with not using any optimization procedure, as well as being robust in case of noisy images, make this method a suitable choice for implementation in everyday used electronic devices; Not to forget parallelizability using GPU based implementations.

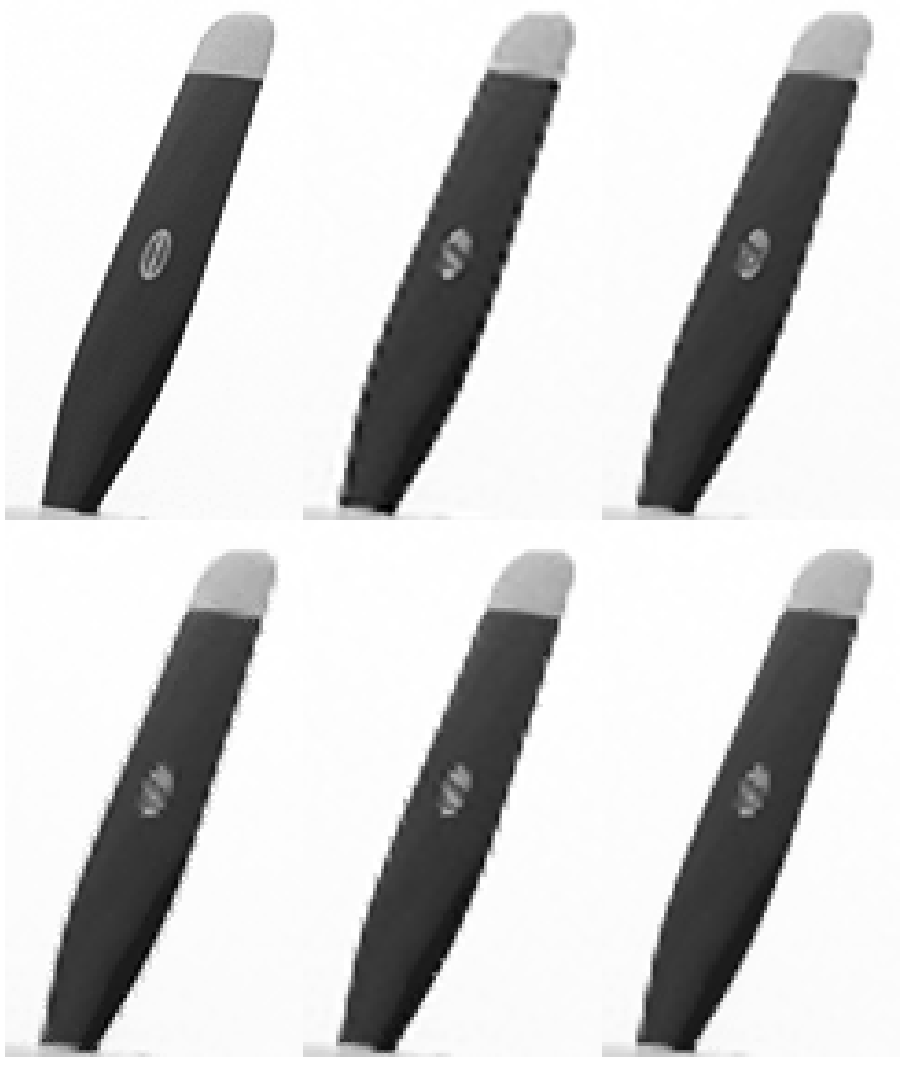

**Fig 7. Subjective quality comparison of different interpolation methods.Top: Original image, ICBI and DFDF. Bottom: NEDI, iNEDI, and STB.**

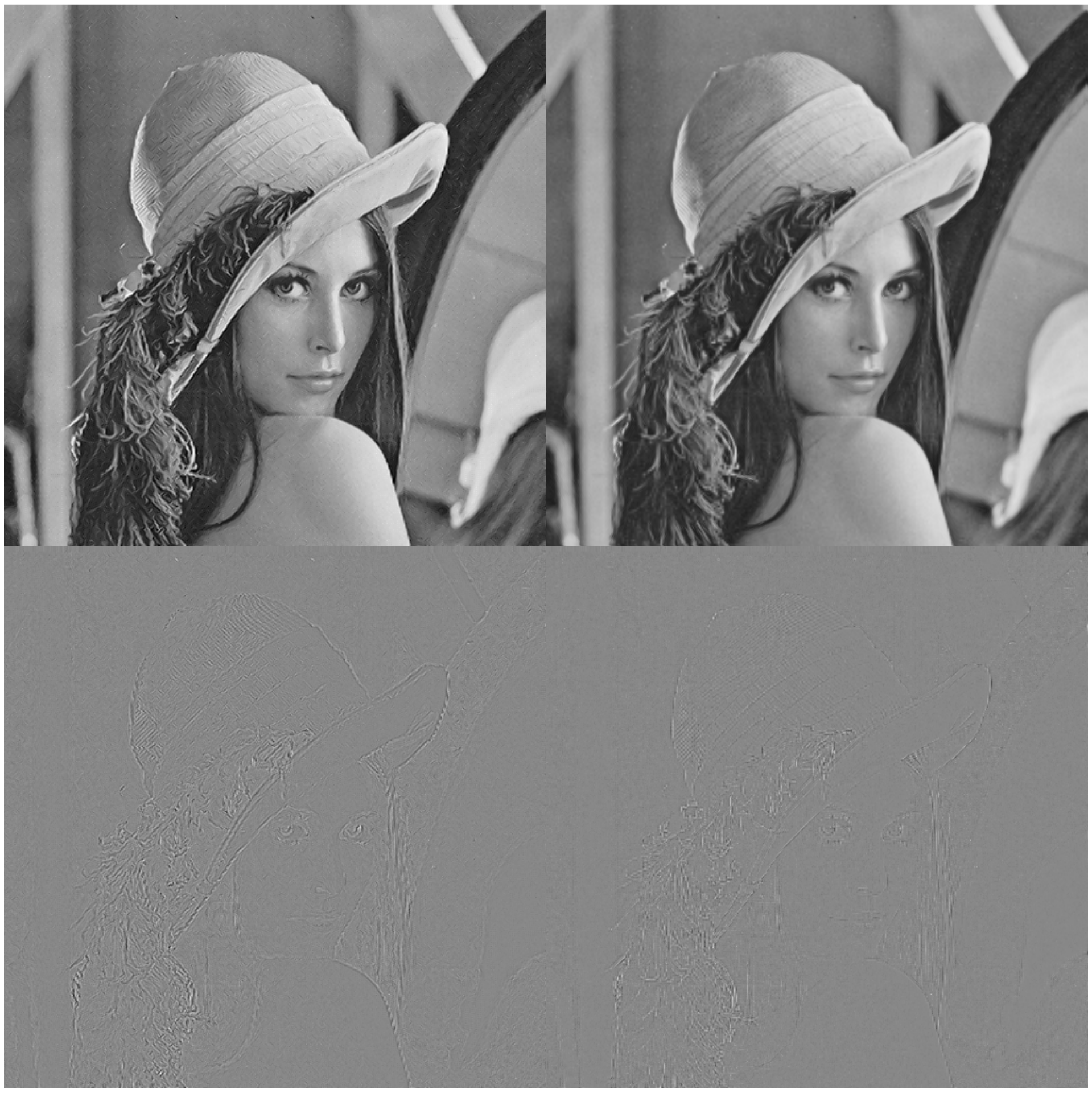

**Fig 8. Subjective quality comparison between the results of RM (PSNR=30.41 , EPSNR=24.89 , SSIM=.8448 , FSIM=.9807) (top left) and STB (PSNR=33.99 , EPSNR=28.70 , SSIM=.9147 , FSIM=.9875) (top right) methods, as well as the difference images with respect to the original Lena image, for RM (bottom left) and STB (bottom right) methods.**